# A support vector regression-based multi-fidelity surrogate model

Maolin Shi[a], Shuo Wang[a], Wei Sun[a], Liye Lv[a], Xueguan Song[a*]

[a] School of Mechanical Engineering, Dalian University of Technology, Linggong Road, Dalian, China, 116024

**Abstract**

Computational simulations with different fidelity have been widely used in engineering design. A high-fidelity (HF) model is generally more accurate but also more time-consuming than an low-fidelity (LF) model. To take advantages of both HF and LF models, multi-fidelity surrogate models that aim to integrate information from both HF and LF models have gained increasing popularity. In this paper, a multi-fidelity surrogate model based on support vector regression named as Co_SVR is developed by combining HF and LF models. In Co_SVR, a kernel function is used to map the map the difference between the HF and LF models. Besides, a heuristic algorithm is used to obtain the optimal parameters of Co_SVR. The proposed Co_SVR is compared with two popular multi-fidelity surrogate models Co_Kriging model, Co_RBF model, and their single-fidelity surrogates through several numerical cases and a pressure vessel design problem. The results show that Co_SVR provides competitive prediction accuracy for numerical cases, and presents a better performance compared with the Co_Kriging and Co_RBF models and single-fidelity surrogate models.

*Keywords:* Multi-fidelity surrogate; Support vector regression; Simulation.

# 1. Introduction

Computer simulations have been widely used to replace the computation-intensive and controlled real-life experiments in engineering design. However, running computer simulations can become computationally prohibitive as is the case of computational fluid dynamics [1]. Indeed, it is still impractical to directly use these simulations with an optimizer to evaluate a lot of design alternatives when exploring the design space for an optimum [2]. This limitation can be addressed by adopting surrogate models, which can build the relationship between the inputs and output of interest based on small numbers of simulations [3]. There are a lot of commonly used surrogate models, such as Polynomial Response Surface (PRS) models [4], Kriging (KRG) [5, 6], Artificial Neural Networks (ANN) [7], Radial Basis Functions (RBF) [8, 9], and Support Vector Regression (SVR) [10]. More detailed illustration and comparison of these techniques can refer to works by Wang et al. [11]. The surrogate modeling techniques mentioned above can accelerate design process, but it is needed to point out that the quality of the surrogate models has considerable impact on the results of the surrogate model-based design optimization. The quality of the surrogate model mainly depends on the size of sample points at which the computer simulations are conducted. It is generally recognized that more sample points can effectively increase the accuracy of a surrogate model but at higher cost [12]. While fewer sample points require lower cost, leading to inaccurate surrogate models. Thus, conflict seems to appear between high accuracy and low cost when building metamodels.

To solve this issue, multi-fidelity surrogate (MFS) models based on high-fidelity (HF) models and low-fidelity (LF) models have been developed in recent years [13-15]. An HF model is one that can accurately describe the properties of the system but with relatively high cost. An LF model is one that can reflect the most prominent characteristics of the system at a considerably less computationally demanding. Thus, more LF samples than HF samples can be obtained at the same computational cost. A general trend of the system can be built from many cheap LF samples and calibrated by a few expensive HF samples, which ultimately results in an MFS model [16]. MFS models have generated considerable recent research interest by virtue of the ability that can effectively combine high accuracy of HF samples with low computational cost of LF samples. Kennedy and O'Hagan established an MFS model using Bayesian approach Gaussian process [17]. Forrester extended the popular method of Kriging to the two-level Co-Kriging model by constructing a correlation matrix containing high fidelity and low fidelity information [18]. Xiao generated multi-level multi-fidelity datasets by using Proper orthogonal decomposition techniques and extended Co-Kriging from two levels to multiple levels on the basis of Forrester's work [19]. Han et improved Co-Kriging by using gradient-enhanced Kriging and a new scaling function and demonstrated that the proposed method was more efficient and accurate in aero-loads prediction [20]. Liu proposed a Kriging based multi-fidelity model composed of a global trend term and a local residual term [21]. Their approach aimed to tackle diverse data structure, e,g, the high fidelity points clustering in some subregions. The above-mentioned MFS models based on Kriging model have become popular and were found to work well. However, the use of Kriging model also induces numerical instability, especially for large size of LF samples, due to covariance matrix inverse operation in the training and prediction of Kriging. In addition to Kriging-based multi-fidelity models, other surrogate-based multi-fidelity models are also attracting widespread interest. Durantin proposed a multi-fidelity surrogate model based on RBF, optimized the parameters by minimizing leave-one-out error [22]. Song reduced computational burden using RBF to approximate discrepancy function directly in MFS model, which avoided optimization of parameters [23]. However, the use of RBF still needs to construct gram matrix which also encounters numerical instability problem.

In this paper, a new support vector regression-based multi-fidelity surrogate model named as Co_SVR is proposed. In Co_SVR, SVR with its outstanding generalization performance is adopted to map the difference between the HF and LF models on the entire domain. Besides, a heuristic algorithm is used to obtain the optimal parameters of Co_SVR. The approximation performance of Co_SVR approach is illustrated using some numerical and engineering cases and a comparison of Co_SVR approach with other single and multi-fidelity surrogate modeling techniques is made. It is expected that more accurate MSF models can be developed with Co_SVR for the same sample HF and LF points.

The remaining of this paper is organized as follows. Section 2 introduces the details of support vector regression and the proposed support vector regression-based multi-fidelity surrogate model. Several numerical examples and an engineering example are given in Section 3 and Section 4 respectively to demonstrate the applicability of the proposed approach, followed by a conclusion and future work in Section 5. The last section concludes this work.

## 2. Proposed approach (Co_SVR)

### 2.1 Support vector regression

SVR is based on support vector machine (SVM) whose purpose is to evaluate the complex relationship between the input and the response of interest through mapping the data into a high-dimensional feature space. Let the $i$-th input be denoted by a dimensional vector, $\boldsymbol{x}_i = (x_{i1}, \ldots, x_{id})$, and its response, $y_i$, respectively. The regression model of SVR can be constituted as follows:

$$y = \boldsymbol{\omega}^T \cdot \varphi(\boldsymbol{x}) + b \tag{1}$$

where $\varphi$ denotes the feature map, $\boldsymbol{\omega}$ is the weight vector and $b$ is the bias term. In SVR, it is necessary to minimize a cost function ($C$) containing a penalized regression error as shown below:

$$Cost = \frac{1}{2}\boldsymbol{\omega}^T \cdot \boldsymbol{\omega} + \frac{1}{2}\gamma \sum_{i=1}^{n} e_i^2 \tag{2}$$

The first part is a weight decay which is used to regularize weight sizes and penalize large weights. The second part is the regression error for all training points. The parameter $\gamma$ determines the relative weight of this part as compared to the first part. Lagrange multipliers method is used to optimize (2) as follows:

$$L(\boldsymbol{\omega}, b, e: \boldsymbol{\alpha}) = \frac{1}{2}\|\boldsymbol{\omega}\|^2 + \gamma \sum_{i=1}^{n} e_i^2 - \sum_{i=1}^{n} \alpha_i \{\boldsymbol{\omega}^T \cdot \varphi(\boldsymbol{x}_i) + b + e_i - y_i\} \tag{3}$$

where $\alpha_i$ are Lagrange multipliers. Through setting the partial first derivatives to zero, the optimum solution can be obtained.

$$\frac{\partial L}{\partial \boldsymbol{\omega}} = 0 \rightarrow \boldsymbol{\omega} = \sum_{i=1}^{n} \alpha_i \varphi(\boldsymbol{x}_i)$$

$$\frac{\partial L}{\partial b} = 0 \rightarrow \sum_{i=1}^{n} \alpha_i = 0$$

$$\frac{\partial L}{\partial e_i} = 0 \rightarrow \alpha_i = \gamma e_i, i = 1, 2, \ldots, n$$

$$\frac{\partial L}{\partial \alpha_i} = 0 \rightarrow \boldsymbol{\omega}^T \cdot \varphi(\boldsymbol{x}_i) + b + e_i - y_i = \gamma e_i, i = 1, 2, \ldots, n$$

Thus,

$$\boldsymbol{\omega} = \sum_{i=1}^{n} \alpha_i \varphi(\boldsymbol{x}_i) = \sum_{i=1}^{n} \gamma e_i \varphi(\boldsymbol{x}_i) \tag{4}$$

where a positive definite kernel is used as follows:

$$K(\boldsymbol{x}_i, \boldsymbol{x}_j) = \varphi(\boldsymbol{x}_i)^T \varphi(\boldsymbol{x}_j) \tag{5}$$

The original regression model in (1) can be modified as follows:

$$\mathbf{y} = \sum_{i=1}^{n} \alpha_i \, \varphi(\mathbf{x}_i)^T \varphi(\mathbf{x}) + b = \sum_{i=1}^{n} \alpha_i \langle \varphi(\mathbf{x}_i)^T, \varphi(\mathbf{x}) \rangle + b \tag{6}$$

For a point of $y_j$ to be evaluated it is:

$$y_j = \sum_{i=1}^{n} \alpha_i \langle \varphi(\mathbf{x}_i)^T, \varphi(\mathbf{x}_j) \rangle + b \tag{7}$$

The $\boldsymbol{\alpha}$ vector can be obtained from solving a set of linear equations:

$$\begin{bmatrix} K + \frac{1}{\gamma} & 1_N \\ 1_N^T & 0 \end{bmatrix} \begin{bmatrix} \alpha \\ b \end{bmatrix} = \begin{bmatrix} y \\ 0 \end{bmatrix} \tag{8}$$

And the solution is:

$$\begin{bmatrix} \alpha \\ b \end{bmatrix} = \begin{bmatrix} K + \frac{1}{\gamma} & 1_N \\ 1_N^T & 0 \end{bmatrix}^{-1} \begin{bmatrix} y \\ 0 \end{bmatrix} \tag{9}$$

## 2.2 Co_SVR

The proposed approach is based on SVR, and its typical form is defined as follows:

$$\mathbf{y} = \sum_{i=1}^{p+q} \alpha_i \langle \varphi(\mathbf{x}_i)^T, \varphi(\mathbf{x}) \rangle + b \tag{10}$$

where $\mathbf{x} = [\mathbf{x_L}, \mathbf{x_H}]^T$, $\mathbf{x_L} = \{x_l^1, x_l^2, \ldots, x_l^p\}$ are the LF training samples, $\mathbf{x_H} = \{x_h^1, x_h^2, \ldots, x_h^q\}$ are the HF training samples, and $\mathbf{y} = [\mathbf{y_L}, \mathbf{y_H}]^T$ are that responses of LF and HF training samples. From Section 2.1, it can be found that the key of SVR-based methods is to design a reasonable kernel function that is able to effectively describe the relationship between the inputs and the outputs. In this paper, the following kernel function is utilized to map the complex relationship between the input and the response of interest into a high-dimensional feature space:

$$\mathbf{K}_{MSF}(x,x) = \varphi(\mathbf{x}_i)^T \varphi(\mathbf{x}_j) = \begin{bmatrix} kernel_{L-L}(\mathbf{x_L}, \mathbf{x_L}) & kernel_{L-H}(\mathbf{x_L}, \mathbf{x_H}) \\ kernel_{H-L}(\mathbf{x_H}, \mathbf{x_L}) & kernel_{H-H}(\mathbf{x_H}, \mathbf{x_H}) \end{bmatrix} (i = 1, \ldots, p+q; j = 1, \ldots, p+q) \tag{11}$$

where $kernel_{L-L}(\mathbf{x_L}, \mathbf{x_L})$, $kernel_{L-H}(\mathbf{x_L}, \mathbf{x_H})$, $kernel_{H-L}(\mathbf{x_H}, \mathbf{x_L})$, and $kernel_{H-H}(\mathbf{x_H}, \mathbf{x_H})$ are defined as follows:

$$kernel_{L-L}(x_l^i, x_l^j) = \sigma_L e^{-\sum_{k=1}^{s} \theta_L^k \left\| x_l^{i,k} - x_l^{j,k} \right\|^2} \quad (i,j = 1, \ldots, p) \tag{12}$$

$$kernel_{L-H}(x_l^i, x_h^j) = \rho^2 \sigma_L e^{-\sum_{k=1}^{s} \theta_L^k \left\| x_l^{i,k} - x_h^{j,k} \right\|^2} \quad (i = 1, \ldots, p; j = 1, \ldots, q) \tag{13}$$

$$kernel_{H-L}(x_h^i, x_l^j) = \rho^2 \sigma_L e^{-\sum_{k=1}^{s} \theta_L^k \left\| x_h^{i,k} - x_l^{j,k} \right\|^2} \quad (i = 1, \ldots, p; j = 1, \ldots, q) \tag{14}$$

$$kernel_{L-L}(x_h^i, x_h^j) = \rho^2 \sigma_L e^{-\sum_{k=1}^{s} \theta_L^k \|x_h^{i,k} - x_h^{j,k}\|^2} + \sigma_d e^{-\sum_{k=1}^{s} \theta_d^k \|x_h^{i,k} - x_h^{j,k}\|^2} \quad (i,j = 1, \ldots, q) \quad (15)$$

where $s$ is the dimension of input. Similar to SVR, the $\boldsymbol{\alpha}$ vector and $b$ in (10) can be obtained with the given $\rho$, $\sigma_L$, $\sigma_d$, $\boldsymbol{\theta}_L$, $\boldsymbol{\theta}_d$, $\boldsymbol{x}$ and $\boldsymbol{y}$ through solving the following linear equations:

$$\begin{bmatrix} K_{MSF} + \frac{1}{\gamma} & 1_N \\ 1_N^T & 0 \end{bmatrix} \begin{bmatrix} \alpha \\ b \end{bmatrix} = \begin{bmatrix} y \\ 0 \end{bmatrix} \quad (16)$$

And the solution is:

$$\begin{bmatrix} \alpha \\ b \end{bmatrix} = \begin{bmatrix} K_{MSF} + \frac{1}{\gamma} & 1_N \\ 1_N^T & 0 \end{bmatrix}^{-1} \begin{bmatrix} y \\ 0 \end{bmatrix} \quad (17)$$

For each point of HF training samples $y_h^j$ to be evaluated it is:

$$\widehat{y_h^j} = \sum_{i=1}^{p+q} \alpha_i \langle \varphi(x_i)^T, \varphi(x_h^j) \rangle + b \quad (18)$$

Thus, the root mean square error ($RMSE$) of HF training samples is used as the cost function of Co_SVR as follows:

$$Cost_{Co\_SVR} = \sqrt{\frac{\sum_{j=1}^{q}(y_h^j - \widehat{y_h^j})^2}{q}} \quad (19)$$

Through minimizing (19), the optimum parameters $\rho$, $\sigma_L$, $\sigma_d$, $\boldsymbol{\theta}_L$, $\boldsymbol{\theta}_d$ are obtained finally.

In the proposed approach, a heuristic algorithm, Grey Wolf optimizer (GWO), is used to obtain the optimum parameters of Co_SVR. The GWO is inspired by the social leadership and hunting technique of grey wolves. In order to mathematical model the social hierarchy of wolves, the first, second and third solutions are considered as the first ($1st$) wolf, the second ($2nd$) and third ($3rd$) wolves, respectively. The rest of the candidate solutions are common ($Common$) wolves. In the GWO algorithm, $1st$, $2nd$ and $3rd$ wolves lead the hunting activities, and the $Common$ wolves follow them in the search for the global optimum. The following equations are introduced to simulate the encircling behavior of grey wolves during hunting:

$$\boldsymbol{D} = |\boldsymbol{C} \cdot \boldsymbol{X}_p(t) - \vec{\boldsymbol{X}}(t)| \quad (20)$$

$$\boldsymbol{X}(t+1) = |\boldsymbol{X}_p(t) - \boldsymbol{A} \cdot \boldsymbol{D}| \quad (21)$$

where $t$ indicates the current iteration; $X_p(t)$ is the position vector of the prey; $X$ is the position vector of a grey wolf which is $[\rho, \sigma_L, \sigma_d, \theta_L, \theta_L]$ in the proposed approach; $A$ and $C$ are coefficient vectors and are calculated as follows:

$$A = 2a \cdot r_1 - a \tag{22}$$

$$C = 2r_2 \tag{23}$$

where elements of $a$ linearly decrease from 2 to 0 over the course of iterations and $r_1$, $r_2$ are random vectors in [0, 1]. To find the optimal solution, the GWO algorithm saves the first three best solutions obtained so far and obliges other candidate solutions to update their positions with respect to them. The following formulas are run constantly for each candidate solution during optimization in order to simulate the hunting and find promising regions of the search space:

$$D_{1st} = |C_1 \cdot X_{1st} - X| \tag{24}$$

$$D_{2nd} = |C_2 \cdot X_{2nd} - X| \tag{25}$$

$$D_{3rd} = |C_3 \cdot X_{3rd} - X| \tag{26}$$

$$X_1 = X_{1st} - A_1 \cdot D_{1st} \tag{27}$$

$$X_2 = X_{1st} - A_2 \cdot D_{2nd} \tag{28}$$

$$X_3 = X_{1st} - A_3 \cdot D_{3rd} \tag{29}$$

$$X(t+1) = \frac{X_1 + X_2 + X_3}{3} \tag{30}$$

The GWO algorithm starts optimization with generating a set of random solutions as the first solutions. The three best obtained solutions so far are saved and considered as $1st$, $2nd$, and $3rd$ solutions. For other solutions, the position is updated through (24) to (30). It is noted that parameters $a$ and $A$ are linearly decreased over the course of iteration. The search agents tend to diverge from the prey when $|A| > 1$ and converge towards the prey when $|A| < 1$. Finally, the position and score of the $1st$ solution is returned as the best optimum parameters of Co_SVR.

2.3 Performance criteria

$R^2$ is selected as the criterion for the performance evaluation, and calculated as follows:

$$R^2 = 1 - \frac{\sum_{i=1}^{n}(y_i - \hat{y}_i)^2}{\sum_{i=1}^{n}(y_i - \bar{y})^2} \tag{31}$$

where $n$ is the number of samples; $y_i$ and $\hat{y}_i$ represent true responses and predictions at testing points, respectively; and $\bar{y}$ is the means of true responses. Essentially, $R^2$ denotes the correlation between the true model and the surrogate model, and the surrogate model is more accurate if $R^2$ is close to one. The Pearson correlation coefficient (PCC), also referred to as $Pearson's\ r$, is a measure of the correlation between two random variables **X** and **Y**. In this paper, we use square of $Pearson's\ r$ which is denoted as $r^2$ to describe the correlation between HF and LF functions as shown follows:

$$r^2 = \left(\frac{\sum_{i=1}^{n}(y_h - \overline{y_h})(y_l - \overline{y_l})}{\sqrt{\sum_{i=1}^{n}(y_h - \overline{y_h})^2}\sqrt{\sum_{i=1}^{n}(y_l - \overline{y_l})^2}}\right) \tag{32}$$

where $y_h$ and $y_l$ denote the HF and LF responses, respectively; $\overline{y_h}$ and $\overline{y_l}$ represent the means of HF and LF responses, respectively.

3. Numerical examples

3.1 Design of experiments

In this section, the performance of Co_SVR is validated and compared with two benchmark MFS models (Co_KRG and Co_RBF) and single-fidelity support vector regression (SVR) through three well-known numerical test problems. For each test problem, the number of HF samples is $2s$, and that of LF samples is $10s$. Design of experiments (DoEs) are the methods to strategically generate samples from computer simulations or experiments in a domain of interest to build surrogate models. Among many available DoE methods, the Latin hypercube sampling (LHS) has been proved to be capable of balancing the trade-off between accuracy and robustness by generating a near-random set of samples. For all surrogate models in this paper, the MATLAB function $lhsdesign$ is adopted to generate DoE samples. To mitigate the impact of random DoE on surrogate performance, 30 sets of DoE samples are generated randomly and the averaged results are compared for the three numerical test problems. In addition, 1000 randomly generated testing points are used for validation.

3.2 Test problem 1: Currin function

In the case of Currin function, the HF and LF models are defined as follows:

HF model:

$$f_H(\pmb{x}) = \left(1 - exp\left(-\frac{1}{2x_2}\right)\right)\left(\frac{2000x_1^3 + 1900x_1^2 + 2092x_1 + 60}{100x_1^3 + 500x_1^2 + 4x_1 + 20}\right) \quad (33)$$

LF model:

$$f_L(\pmb{x}) = (1 - m^2 - 2m)f(x_1 + 0.05, x_2 + 0.05) + \frac{1}{4}(f(x_1 + 0.05, \max(0, x_2 - 0.05)) + f$$

$$(x_1 - 0.05, x_2 + 0.05) + f(x_1 - 0.05, \max(0, x_2 - 0.05))) \quad (34)$$

where $\pmb{x} \in [0, 0.5]$, $f_H(\pmb{x})$ is an HF model, $f_L(\pmb{x})$ is an LF model, and the parameter $m$ varies from 0 to 1 to reflect the degree of the correlation $r^2$.

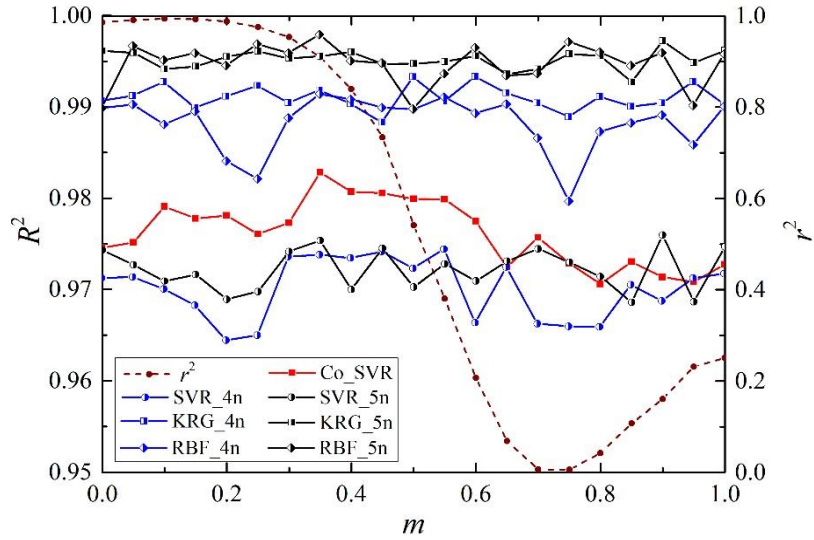

Figure 1 Comparison between MFS-RBF and single-fidelity surrogate models

Figure 1 compares the MFS-RBF model with single-fidelity surrogate models. Two sample sets, 4n and 5n, are generated to construct different single-fidelity surrogate models. To eliminate the effect of DoE, the accuracies of the single-fidelity surrogates are averaged over 30 randomly sampling sets. The red dashed line in Fig. 1 shows the relationship between correlation $r^2$ and the parameter $m$ for the Currin function. It is observed that the minimum $r^2$ value is obtained when $m = 0.75$. The $r^2$ between the HF and LF models decreases as $m$ increases from 0 to 0.75, while the $r^2$ increases as $m$ increases from 0.75 to 1. A maximum $r^2$ is obtained when $m = 0.1$. It is seen from Fig. 1 that the tendency for the performance of Co_SVR matches the tendency of the correlation $r^2$. From Fig. 1, we can see that Co_SVR with $2s$ HF samples almost outperforms single-fidelity SVR no matter the sample number is $4s$ or $5s$ but produces smaller $R^2$s than single-fidelity KRG and RBF. Considering the mean $R^2$s of Co_SVR are all higher than 0.97, its prediction performance is still acceptable.

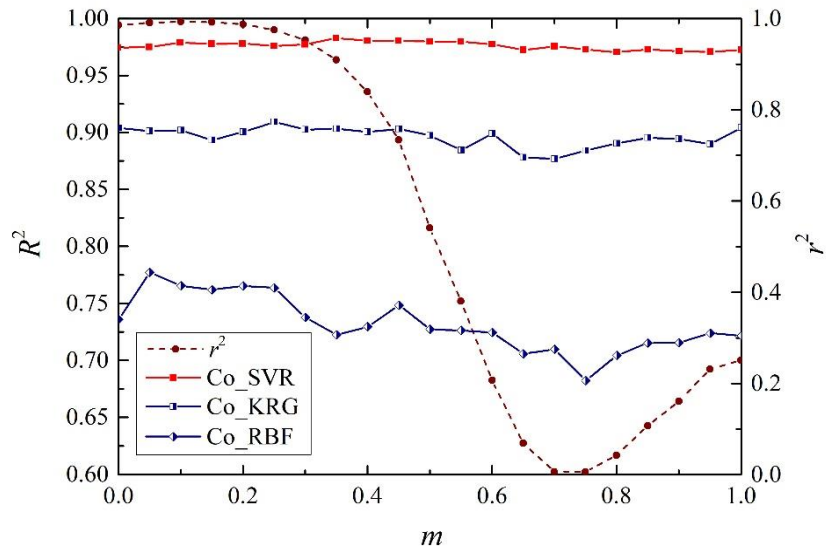

Figure 2 Comparison between MFS-RBF and single-fidelity surrogate models

Figs. 2&3 compare the Co_SVR, Co_KRG, and Co_RBF models with $2s$ HF samples and $10s$ LF samples on $R^2$ and the standard deviation ($Std.$) of $R^2$. Each value of $R^2$ at parameter $m$ is the average of the results obtained for 30 DoEs, and the $Std.$ of $R^2$ denotes the standard deviation of the 30 values. The results show that MFS-RBF performs much better than Co_KRG and Co_RBF. It is found that the tendency of the performance of the Co_SVR, Co_KRG, and Co_RBF models as shown in Fig. 2 is consistent with the tendency of the HF/LF model correlation $r^2$ as shown in Fig. 1. In addition, it is seen that the Co_SVR model shows a larger $R^2$ and a smaller $Std.$ of $R^2$, which performs better than Co_KRG and Co_RBF models in terms of both prediction accuracy and robustness.

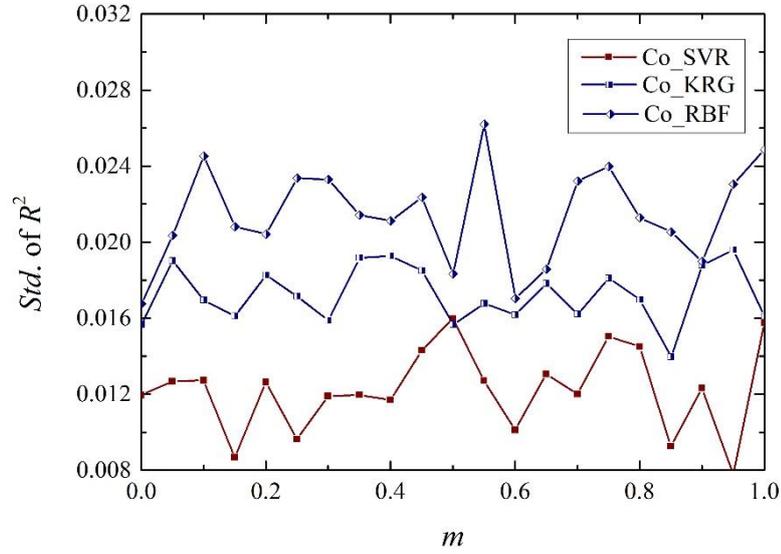

Figure 3 Comparison between MFS-RBF and single-fidelity surrogate models

3.3 Test problem 2: Park function 1

The HF and LF Park function 1 is defined as follows:

HF model:

$$f_H(x) = \frac{x_1}{2}\left[\sqrt{1 + (x_2 + x_3^2)\frac{x_4}{x_1^2}} - 1\right] + (x_1 + 3x_4)exp[1 + sin(x_3)] \qquad (35)$$

LF model:

$$f_L(x) = \left[1 + \frac{sin(x_1)}{10}\right]f_H(x) - 2x_1 + x_2^2 + x_3^2 + 0.5 \qquad (36)$$

where $x \in [-1, 0]$. To study the effect of the HF/LF model correlation on the performance of Co_SVR, the LF function in (36) is changed by multiplying a coefficient function of parameter $m$ with its first term as follows:

$$f_L(x) = (1 - m^2 - 2m)\left[1 + \frac{sin(x_1)}{10}\right]f_H(x) - 2x_1 + x_2^2 + x_3^2 + 0.5 \qquad (37)$$

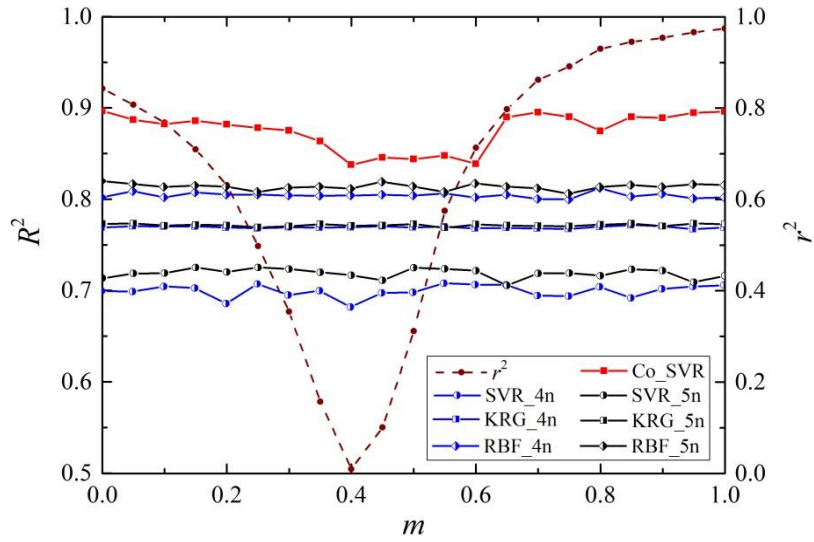

Figure 4 Comparison between MFS-RBF and single-fidelity surrogate models

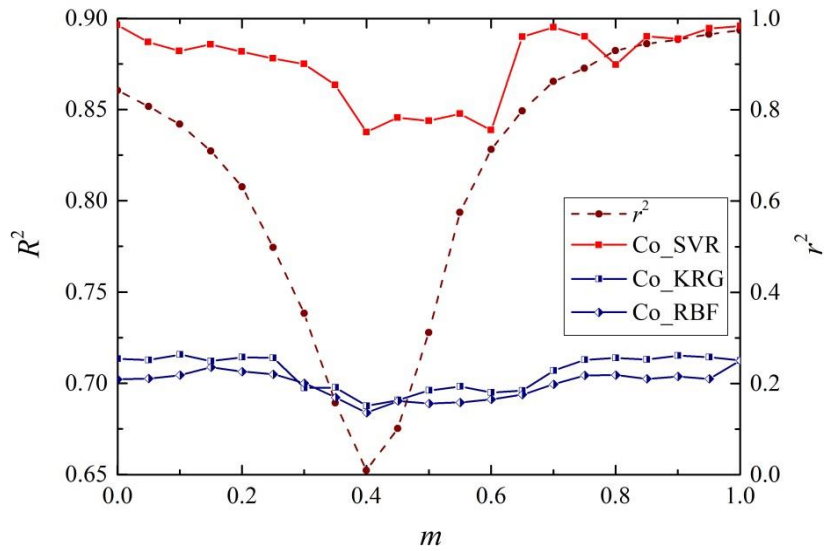

Figure 5 Comparison between MFS-RBF models

Figure 4 compares the Co_SVR model with single-fidelity surrogate models. Two sample sets, $4s$ and $5s$, are generated to construct different single-fidelity surrogate models. From this figure, it can be found that the minimum $r^2$ is obtained when $m = 0.4$. The $r^2$ between the HF and LF models decreases as $m$ increases from 0 to 0.4, while the $r^2$ increases as $m$ increases from 0.4 to 1. A maximum $r^2$ is obtained when $m$ is 1. The tendency for the prediction accuracy of Co_SVR matches the tendency of the correlation $r^2$. The Co_SVR with $2s$ HF samples outperforms all single-fidelity surrogates no matter the sample number is 4n or 5n as $m$ varies from 0 to 1. Figure 5 compares the Co_SVR, Co_KRG, and Co_RBF models with 2s HF samples and 10s LF samples on $R^2$. Each value of $R^2$ at parameter $m$ is the average of the results obtained for 30 DoEs. From this figure, it can be found that the Co_SVR model outperforms the Co_KRG, and Co_RBF models as $m$ varies from 0 to 1. It is interesting to find that the tendency of the performance of the Co_KRG, and Co_RBF models is consistent with the tendency of the HF/LF model correlation $r^2$, while the LR-MFS model is relatively insensitive to the correlation $r^2$. Figure 6 illustrates the standard deviation of $R^2$. It is seen that the Co_SVR model produces a larger standard deviation of $R^2$ than the Co_KRG, and Co_RBF models when $m$ is smaller than 0.55. When $m$ varies from 0.55 to 1, the results of the Co_SVR model is similar to the Co_KRG, and Co_RBF models. Student's test is used to statistically compare the performance of the Co_SVR model with the Co_KRG and Co_RBF models based on the mean and standard deviation of $R^2$. The null hypothesis is the performance of the Co_SVR model is better than the Co_KRG/Co_RBF models. The results of Student's test is shown in Figure 7. From this figure, it can be found that the statistics $t$ is much higher than $t_{0.95/58} = 1.65$ at different $m$. The Co_SVR model performs significantly better than Co_KRG and Co_RBF models for Park function 1.

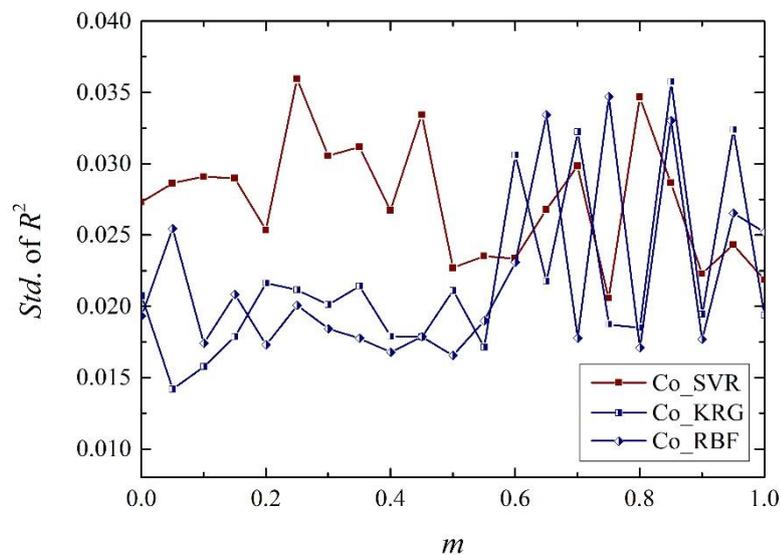

Figure 6 Comparison between MFS-RBF models

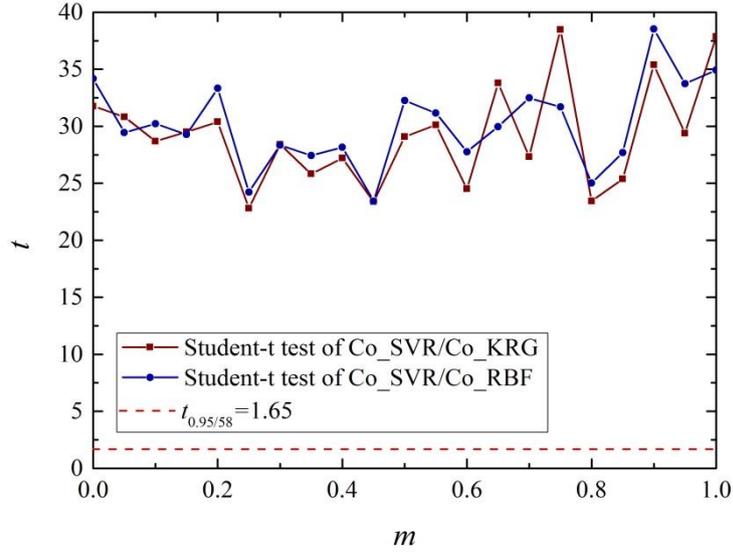

Figure 7 Student's tests between MFS-RBF models

### 3.4 Test problem 3: Park function 2

The HF and LF Park function 2 are defined as follows:

HF model:

$$f_H(x) = \frac{2}{3}exp(x_1 + x_2) - x_4 sin(x_3) + x_3 \tag{38}$$

LF model:

$$f_L(x) = 1.2 f_H(x) - (0.5m^2 + m + 0.5) * \frac{2}{3}exp(x_1 + x_2) \tag{39}$$

where $x \in [0, 1]$. Figure 5 compares the MFS-RBF model with single-fidelity surrogate models. Two sample sets, 8n and 10n, are generated to construct different single-fidelity surrogate models. It is observed that the least $r^2$ occurs when $m = 0.68$. When A $\leq 0.68$, $r^2$ monotonically decreases from 0.61 to 0. When $m \geq 0.2$, $r^2$ monotonically increases from 0 to 1. It is seen that the tendency of the Co_SVR performance strongly matches the tendency of HF/LF correlation $r^2$. When the correlation $r^2$ is higher than 0.3, namely $m$ is in $[0, 0.1] \cup [0.25, 1]$, the Co_SVR model performs the best. When the correlation $r^2$ is less than 0.3, namely $0.1 \leq m \leq 0.25$, the Co_SVR model performs better than most single surrogate models except "RBF_4n" and "RBF_5n".

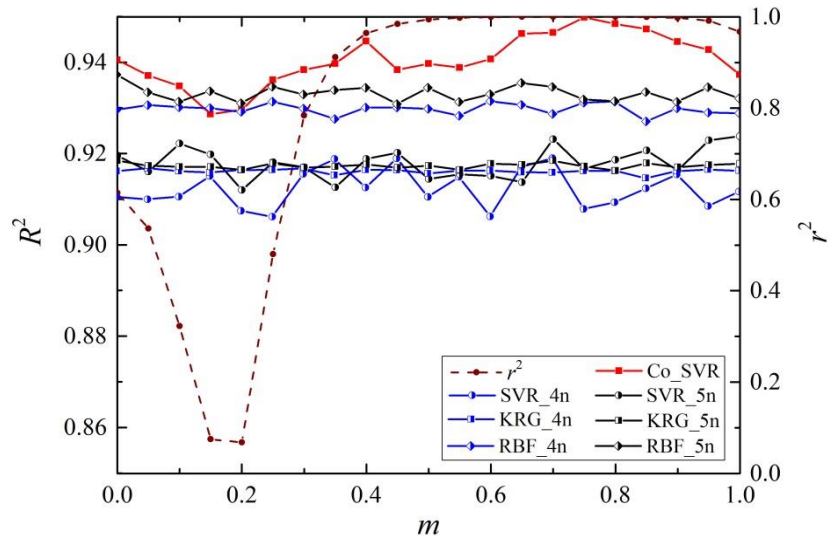

Figure 8 Comparison between MFS-RBF and single-fidelity surrogate models

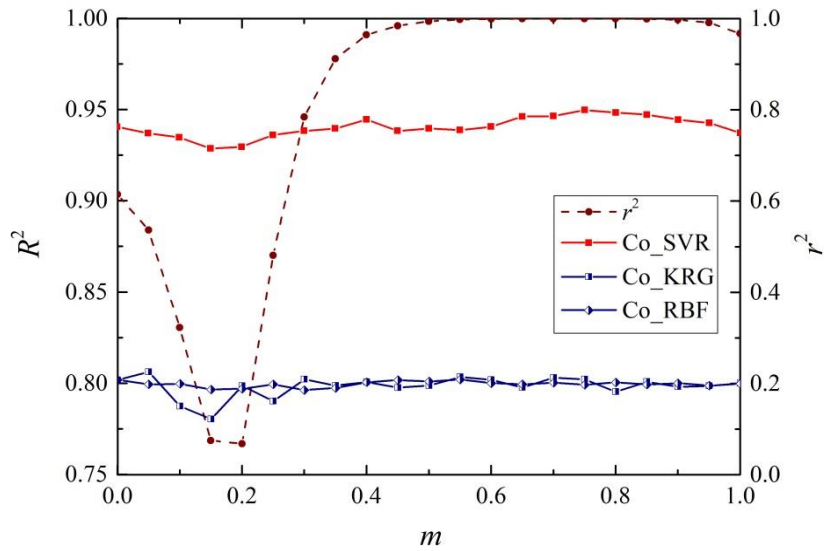

Figure 9 Comparison between MFS-RBF models

Figures 9&10 compares the Co_SVR, Co_KRG, and Co_RBF models based on $R^2$ and $Std.$ of $R^2$. Form these two figures, it can be found that the Co_SVR model provides much better prediction accuracy, and produces similar $Std.$ of $R^2$ to the other two baselines MFS models. The accuracy performance of Co_SVR is competitive for Park function 2 compared with other MFS models.

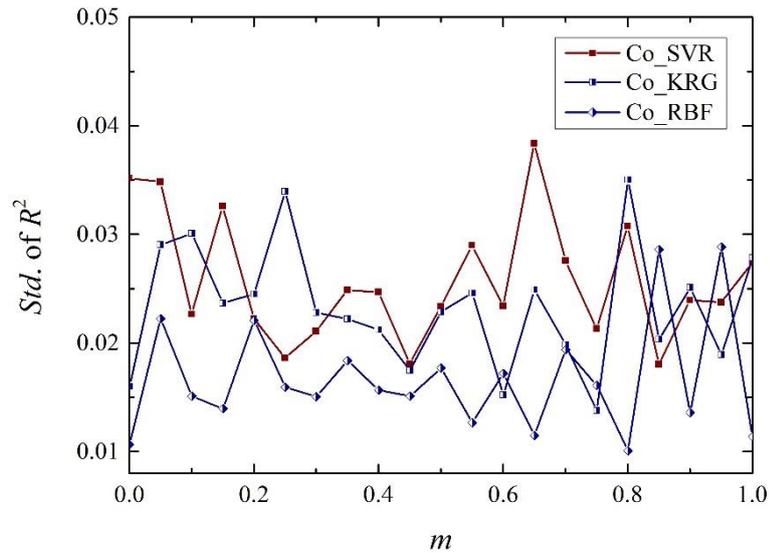

Figure 10 Comparison between MFS-RBF models

## 4. Engineering problem

In addition to the four numerical problems, there is an engineering problem, computational fluid dynamics (CFD) analysis of a pressure relief valve (PRV), to validate the performance of the proposed Co_SVR. A PRV is a type of safety valves used to provide overpressure protection in many engineering industries such as petrochemical, natural gas processing, and power generation industries. In general, PRVs are connected to pressure vessels, boilers or other equipment via a piping system. The operation of the PRV is based on the balance of force. As shown in Figure 11a, when the resultant force which mainly consists of fluid force ($F$) and spring force is going upward, the disc is open and a portion of the fluid from inlet passes the gap between the seat and disc, then escapes through the outlet of the valve. Thus, the pressure vessels or other equipment are protected from overpressure. After part of the fluid is discharged, the fluid force ($F$) may declines. Once the resultant force drops to zero, the valve will close. If the valve is not well designed, the valve may open very slowly without popping, it thus cannot release enough fluid quickly to control the over-pressure; on the other hand, if a valve takes a long time to reclose, it may release excessive amounts fluid, besides resulting in product lost or atmospheric contamination, and it also lead to unsteady pressure for the upstream or the downstream system. The fluid force is an important factor to consider when designing a PRV. In order to obtain the fluid force (*F*), two steady simulations of CFD with different dimensions are performed using commercial software Fluent. The standard $k-\varepsilon$ turbulence model is employed; the medium is water with an initial temperature of 300 K. Three-dimensional (3-D) CFD model including 284, 412 unstructured mesh (Fig. 11b) and two-dimensional axisymmetric CFD model including 9646 unstructured grid (Fig. 11c) are used as the HF and LF simulation models, respectively. In these two kinds of CFD simulations, the pressure of inlet was set to constant and the pressure of outlet was zero. The pressure of outlet (*P*) and opening lift (*L*) were selected as two design variables ranging from 0.1 to 0.4 atm and 1 to 4 mm, respectively.

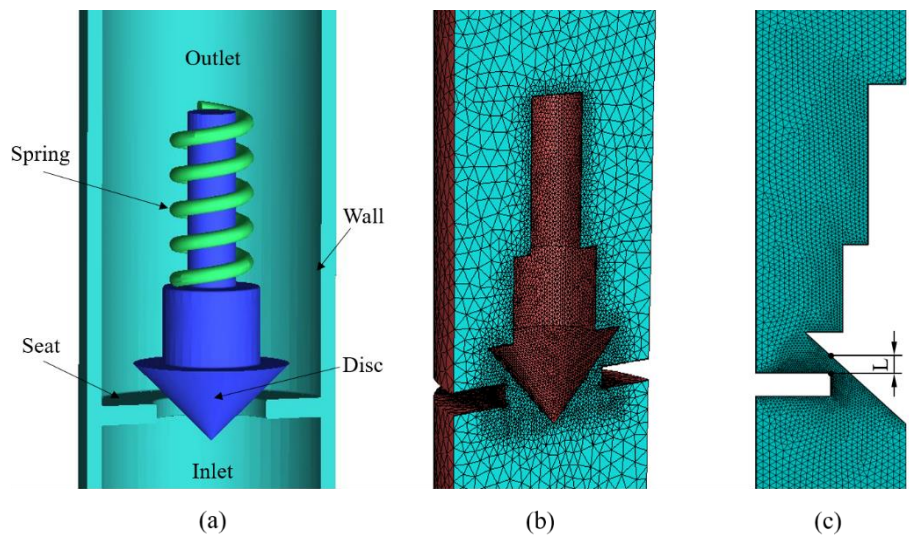

Figure 11 Comparison between MFS-RBF and single-fidelity surrogate models

In this section, a set consisting of 20 samples is generated and used as the training set, and another set consisting of 20 samples is used as the testing set. For each point of the training set, the HF and LF simulations are both conducted. It is found that running one HF simulation takes about 50min while running one LF simulation takes about 10 min on a computer configured with an Intel Core i7 6700 CPU and 16G RAM. The accuracy of the MFS model was verified by cross-validation (CV). The HF training samples were divided randomly into 5 sets, i.e., each set contains 4 HF samples. In each experiment, totally 20 LF training samples and one set of HF samples are used to construct the MFS models, and the same set of HF samples is used to construct the single-fidelity models. The accuracy of the MFS models and single-fidelity models by the testing samples. This process was repeated 5 times. Eventually, the accuracy was averaged out. The comparison of Co_SVR with the other two MFS models and three single-fidelity surrogate models are shown in Figure 12. It is seen that Co_SVR performs best compared with the other techniques in terms of the engineering problem. In addition, the three MFS models all outperforms their single-fidelity models, respectively, which indicates the MSF model is able to improve the prediction accuracy by the help of LF samples.

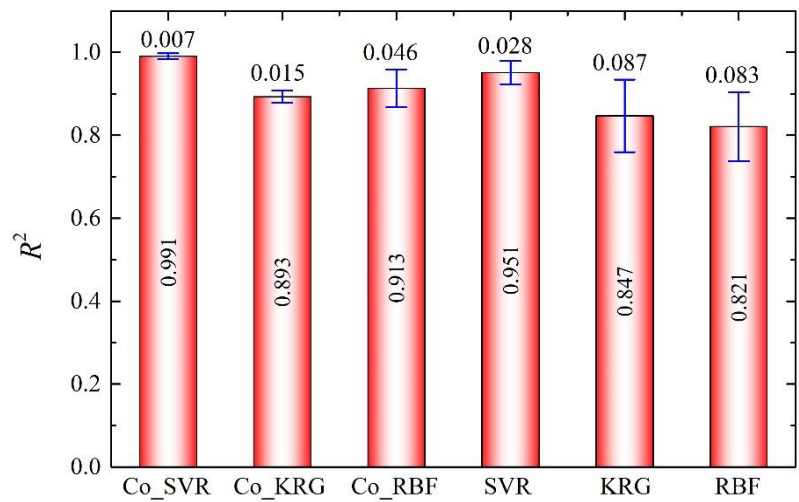

Figure 12 Comparison of different MFS and single-fidelity models for engineering problem

## 5. Conclusions

In this paper, we proposed new multi-fidelity model based on support vector regression. A special kernel function is used in the proposed Co_SVR to map the relationship between the HF and LF models on the entire domain. Besides, a heuristic algorithm is used to accelerate the train process of Co_SVR. The proposed approach is compared to five different metamodeling methods (Co_KRG, Co_RBF, SVR, KRG and RBF) using several numerical cases and an engineering design problem under different correlations between HF and LF models. It is concluded that the proposed Co_SVR using less HF samples exhibits competitive performance compared with single-fidelity surrogate models. Co_SVR also performs better than the other two MFS models for both numerical cases and engineering cases.

In engineering practices, the designers usually use relatively low-fidelity simulations to study the system behavior, then gradually increase the fidelity of simulations to accurately describe the physical features of the system. Multiple fidelity samples (generally more than two) can be obtained in the previous design process. The proposed Co_SVR only used the HF samples and the samples with same low-fidelity, but the samples with other low-fidelity are not utilized. Thus, extending the Co_SVR to solve engineering design with multiple low-fidelity samples and high-fidelity samples will be investigated in our future work. Overall, as a novel variable-fidelity modeling technique, Co_SVR exhibits great capability for simulation based engineering design and optimization problems.

## Acknowledgments

The research is supported by the National Natural Science Foundation of China (Grant No. U1608256).